\documentclass[final]{cvpr}

\usepackage{times}
\usepackage{epsfig}
\usepackage{graphicx}
\usepackage{amsmath}
\usepackage{amssymb}
\usepackage{booktabs}
\usepackage{multirow}
\usepackage{url}

\usepackage{subcaption}
\newcommand\norm[1]{\left\lVert#1\right\rVert}

\usepackage[pagebackref=true,breaklinks=true,colorlinks,bookmarks=false]{hyperref}
\usepackage{multirow}

\def\cvprPaperID{****} 
\def\confYear{CVPR 2021}

\begin{document}

\title{Simple Distillation Baselines for Improving Small Self-supervised Models}

\author{Jindong Gu\\
University of Munich\\
\and
Wei Liu\\
Tencent\\
\and
Yonglong Tian\\
MIT
}

\maketitle

\begin{abstract}
While large self-supervised models have rivalled the performance of their supervised counterparts, small models still struggle. In this report, we explore simple baselines for improving small self-supervised models via distillation, called SimDis. Specifically, we present an offline-distillation baseline, which establishes a new state-of-the-art, and an online-distillation baseline, which achieves similar performance with minimal computational overhead. We hope these baselines will provide useful experience for relevant future research. Code is available at: 
\small{\url{https://github.com/JindongGu/SimDis/}}

\end{abstract}

\section{Introduction}

Recent self-supervised learning algorithms have largely closed the gap in linear classifier between unsupervised and supervised representations from large models (\eg, the gap is only $0.5\%$ for BYOL using ResNet-200x2 on ImageNet)~\cite{grill2020bootstrap}. However, such gap increases as we reduces the model capacity. For instance, the linear probing accuracy of BYOL using ResNet-50 is $2.3\%$ lower than its supervised counterpart. With a smaller model such as ResNet-18, this gap even increases to $8.9\%$ (our reimplementation).

We speculate that smaller models with limited representation power cannot directly solve the self-supervised pre-training task well. This is because such task is often difficult, \eg, contrastive learning requires the model to recognize each instance in a large scale dataset. Meanwhile, small models are usually preferred to run for real-world applications, as it reduces latency, power cost, and carbon emission. Therefore, it is of significant importance to learn small but powerful self-supervised models.

\begin{figure}[t]
    \centering
    \includegraphics[scale=0.5]{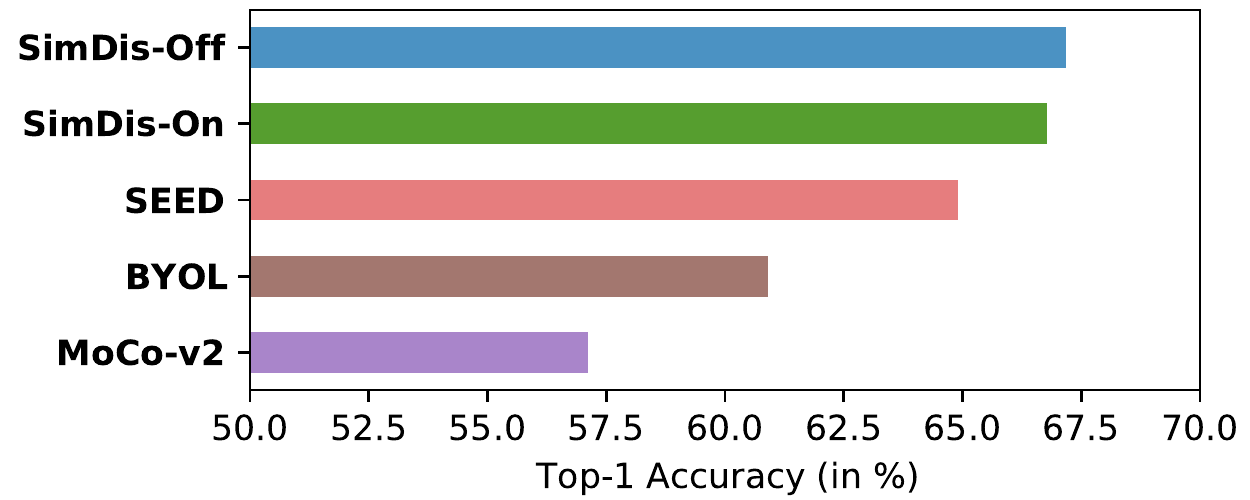}
    \caption{Linear evaluation accuracy of self-supervised ResNet18: Our baselines SimDis achieve SOTA performance.}
    \label{fig:teaser}
\end{figure}

In this report, we build two strong baselines (called \emph{SimDis}) for improving small self-supervised models, by exploring simple distillation strategies. Distillation has been successfully used to compress large models in supervised setup but rarely considered for self-supervised learning. Our first baseline is offline: we firstly train a large model until convergence, and then distill to a small model. This baseline significantly outperforms previous state-of-the-art SEED~\cite{fang2021seed}, without using memory buffers. For the second baseline, we simultaneously train the small model with the large model by exploiting various outputs from the large model as teaching \emph{views}. With minimal computational cost, this online baseline (called \emph{SimDis-On}) achieves similar performance as the offline baseline (called \emph{SimDis-Off}), as shown in Figure~\ref{fig:teaser}. Moreover, we demonstrate that these two baselines work well when LARS and Synchronized BatchNorm (both are critical for many self-supervised learning algorithms) are removed.

\begin{figure*}[t]
    \centering
        \includegraphics[width=0.75\textwidth]{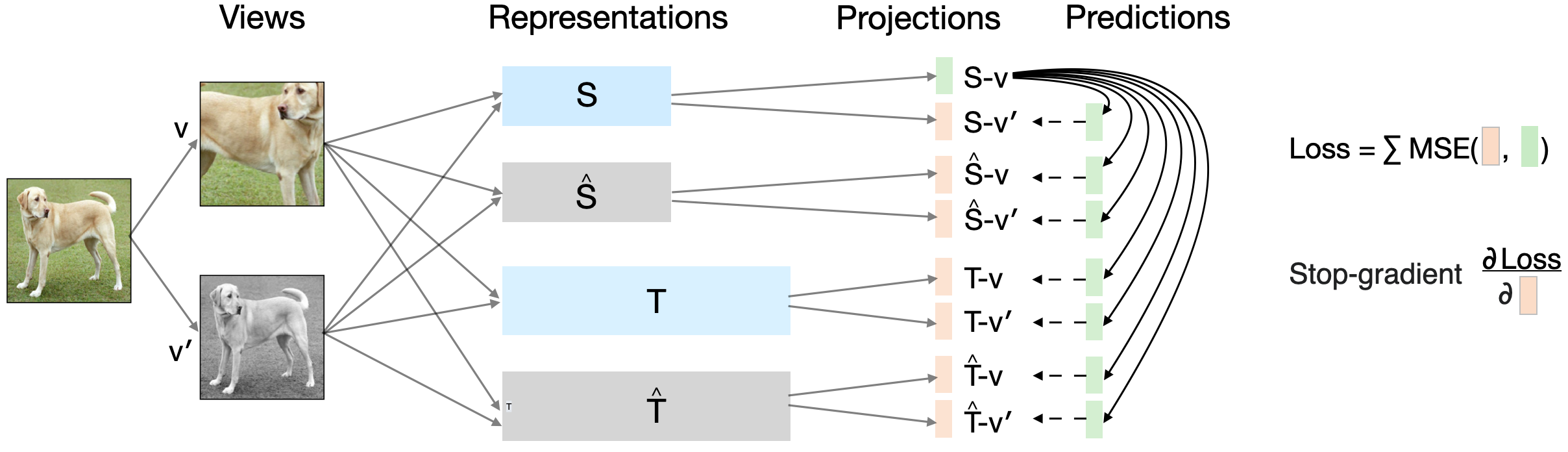}
        \caption{The overview of our distillation approach: The model consists of a teacher (a siamese network marked with $T$ and $\hat{T}$) and a student model (a siamese network marked with $S$ and $\hat{S}$). The student $S$ is updated to predict projected representations of both the student and the teacher.}
        \label{fig:overview}
\end{figure*}

\section{Related Works}
\noindent \textbf{Self-supervised Learning.} Recently, self-supervised learning has been significantly advanced by contrastive learning, which either takes explicit negatives~\cite{oord2018representation,wu2018unsupervised,tian2019contrastive,he2020momentum,chen2020simple} or implicit regularization~\cite{grill2020bootstrap,caron2020unsupervised}. However, contrastive learning is desperate for model size, \ie, the larger the model is, the smaller the gap between the accuracies of self-supervised and supervised representations. Thereore, in this work we focuses on improving small models.

\noindent \textbf{Knowledge Distillation.} In supervised learning setting, knowledge distillation has been broadly studied from different perspectives, e.g., transferring knowledge~\cite{hinton2015distilling,ba2013deep,romero2014fitnets,you2017learning}, model architectures~\cite{gu2020search}, and distillation scheme~\cite{hinton2015distilling,zhang2018deep}. Recent work CompRess~\cite{koohpayegani2020compress} and SEED~\cite{fang2021seed} construct label-free knowledge to guide the training of small self-supervised models. Different from their work, our work provides baselines that are simpler (\eg, removing memory back and negative samples), stronger, and computationally efficient.

\section{Self-supervised Model Distillation}
\label{sec:app}
The overview of our distillation approach is shown in Fig.~\ref{fig:overview}, which consists of a small student model and a big teacher model. Both the student and the teacher contains two siamese networks, which are commonly used in the self-supervised learning paradigm. 
Therein, the blue-grey pair (blue indicates the online network while gray refers to the momentum network) with the bigger rectangles corresponds to the teacher, while the smaller ones to the student. At the end of the training, we only keep the network marked with $S$ as the final small backbone. 

\begin{table*}[h]
\begin{center}
\begin{tabular}{c|c|c|ccc} 
\toprule
\multirow{2}{*}{Methods} & \multirow{2}{*}{\# of Views} & \multirow{2}{*}{FLOPs} & \multicolumn{3}{c}{Training Epochs} \\ 
 &  &  & $N$=100 & $N$=300 & $N$=1000 \\ 

\midrule
SimDis-Off & 2 & (8.2G + 1.8G) $\times M\times N$ & \textbf{61.76} & \textbf{65.15} & \textbf{67.18} \\
SimDis-On & 2 & (4.1G + 1.8G) $\times M\times N$  & 58.08 & 64.49 & 66.78 \\ 
SimDis-On-7v & 7 & (4.1G + 1.8G + 12M) $\times M\times N$ & 60.65 & 64.58 & 66.14\\ 
\bottomrule
\end{tabular}
\caption{The FLOPs and top-1 accuracy (in \%) of each method is reported. $M, N$ stands for the number of training examples and the number of training epochs. 4.1G corresponds to the FLOPs of forward inference of a single image on $T$, 1.8G to the ones on $S$, and the 12M brought by multi-view predictions can be ignored. Given limited training time budgets, the offline method outperforms online one. The performance gap can be reduced by multi-view distillation. When the models are trained for more epochs, they all performs similarly.}
\label{tab:on_vs_off}
\end{center}
\end{table*}

\noindent \textbf{Teacher.} The teacher model with the siamese network architecture can be trained with various self-supervised learning methods. In this work, the recent state-of-the-art method BYOL~\cite{grill2020bootstrap} is applied. Similar to other methods, BYOL uses two neural networks: the online network (the blue rectangle marked with ${T}$) and the target network (the gray rectangle marked with $\hat{T}$).  The online network T is defined by a set of weights $\theta$ and is comprised of three stages: an encoder $f^T_{\theta}$, a projector $g^T_{\theta}$ and a predictor $q^T_{\theta}$.  The target network $\hat{T}$ with the parameters $\xi$ is an exponential moving average of ${T}$.  

Given an image $x$, BYOL produces two augmented views $v$ and $v'$. The representation, the projection and the prediction of $v$ are $y^T_{\theta}= f^T_{\theta}(v)$, $z^T_{\theta}= g^T_{\theta}(y^T_{\theta})$ and $q^T_{\theta}(z^T_{\theta})$, respectively. The training loss measures the mismatch between the prediction of $v$ on T and the projection of $v'$ on $\hat{T}$, namely, the mean squared error between the normalized predictions and projections,
\begin{equation}
\mathcal{L}^T_{\theta,\xi} = \norm{\overline{q^T_{\theta}}(z^T_{\theta})-\overline{{z'}^{\hat{T}}_{\theta}}}^2_2,
\label{equ:loss_T}
\end{equation}
where the overline symbol means $l_2$-normalized. Symmetrically, the mismatch between the prediction of $v'$ on ${T}$ and the projection of $v$ on $\hat{T}$ is also taken as part of the final loss.

At each training step, the parameters $\theta$ of $T$ is updated with gradient-based optimizer. Note that the gradients received by $\hat{T}$ are stopped during optimization. The parameters $\xi$ of $\hat{T}$ is updated with moment-based update rule:
\begin{equation}
\xi \leftarrow \tau \xi + (1- \tau ) \theta,
\label{equ:Tema}
\end{equation}
where $\tau \in[0, 1]$ is specified as the target decay rate.

\noindent \textbf{Student.} The student model is a similar siamese architecture as the teacher but uses a smaller encoder $f^S_{\theta}$. 

The loss used to update the student includes two parts: BYOL loss and Distillation loss. Similar to the loss in Eq.~\ref{equ:loss_T}, the BYOL loss measures the mean squared error between the normalized predictions on $S$ and projections on $\hat{S}$. The Distillation loss measures the mismatch between the predictions on $S$ and projections of another $\hat{T}$ of the same image. Note that two prediction heads are built in the student, one for predicting the projections on $\hat{S}$ and another for predicting the projections on $\hat{T}$. $\hat{T}$ is a stable version of $T$ during training, while they perform similarly at the end of the training. Hence, $\hat{T}$ is chosen across the paper. See Sec.~\ref{sec:vs} for more design choices.

The updates of the student are similar to the ones of the teacher. Namely, the parameters of $S$ are updated with SGD, while the parameters of $\hat{S}$ is updated with moment-based update rules as in Eq.~\ref{equ:Tema}. The only difference is the distillation loss term, which guides the training of the student with the representations learned by the teacher. The updates of the student and the teacher can be simultaneous in a single stage or separate in two stages, which correspond to the online and offline distillation schemes.

\subsection{Offline/Online Distillation Schemes}
\noindent\textbf{Offline.} The offline distillation scheme consists of two stages. In the first stage, only the teacher model is trained as described in Sec.~\ref{sec:app}. In the second stage, the pre-trained teacher model is fixed, and the student model is trained using both BYOL and Distillation losses.  This method is dubbed as \emph{SimDis-Off}.

\noindent\textbf{Online.} The online distillation scheme updates the student and the teacher simultaneously. The teacher model is trained with BYOL loss, which is the same as the first stage of SimDis-Off. At the same time, the student is updated to minimize the BYOL loss and the Distillation loss. Correspondingly, this method is termed as \emph{SimDis-On}.

\subsection{Online vs. Offline} 
\label{sec:vs}
\noindent\textbf{Setting.} The online distillation train both the student and the teacher for $N$ epochs, while the offline one first trains the teacher for $N$ epochs and then the student for $N$ epochs.

\noindent\textbf{Pros and Cons.} In this setting, compared to online distillation, the offline one requires more computational resources, namely the extra cost to do forward inference on the whole training dataset for N times (See Tab.~\ref{tab:on_vs_off}). Note that the extra cost cannot be saved by caching the representations because the augmented images vary with epoch.

The online distillation has also limitations. When the training epoch $N$ is small, the teacher is not well trained at the beginning of the training; the representations extracted by the teacher are not very helpful for guiding the training of the student (See $N=100$ in Tab.~\ref{tab:on_vs_off}). 

\noindent \textbf{Improvement.} To overcome the limitation, we propose multi-view distillation. Specifically, we replace the single-view distillation term with the multi-view distillation loss. The new loss computes the MSE between 6 predictions and the corresponding projections. The projections to predict are $\overline{{z'}^{S}_{\theta}}, \overline{{z}^{\hat{S}}_{\theta}}, \overline{{z}^{T}_{\theta}}, \overline{{z'}^{T}_{\theta}}, \overline{{z}^{\hat{T}}_{\theta}},$ and $\overline{{z'}^{\hat{T}}_{\theta}}$. The projection $\overline{{z'}^{\hat{S}}_{\beta}}$ is already computed in the first BYOL loss term. There are 7 views in total to predict. Hence, this method is dubbed as \emph{SimDis-On-7v}. Although multiple projections (views) are required to compute loss, only tiny extra computational cost is required since the 6 views are free to use in online case. 

\begin{figure*}[h!]
    \centering
    \begin{subfigure}[b]{0.48\textwidth}
        \includegraphics[width=\textwidth]{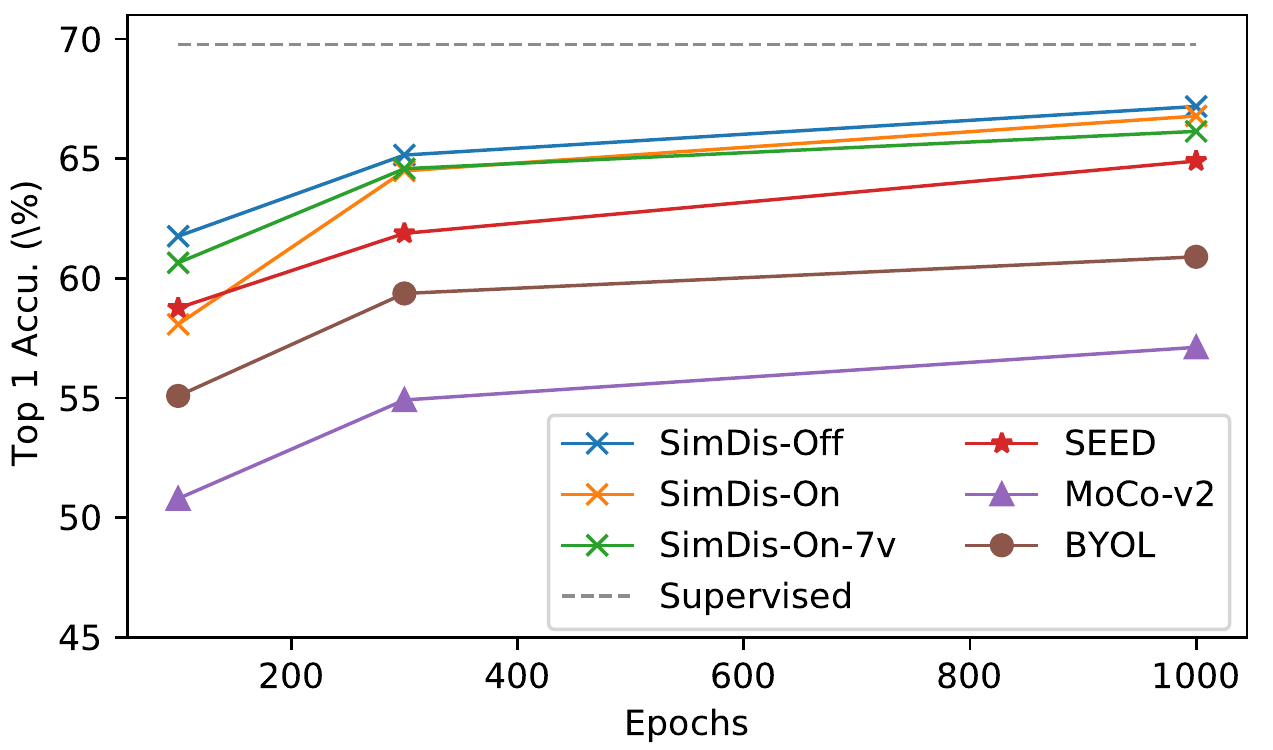}
        \caption{Top-1 Accuracy}
        \label{fig:top1}
    \end{subfigure} \hspace{0.3cm}
     \begin{subfigure}[b]{0.48\textwidth}
        \includegraphics[width=\textwidth]{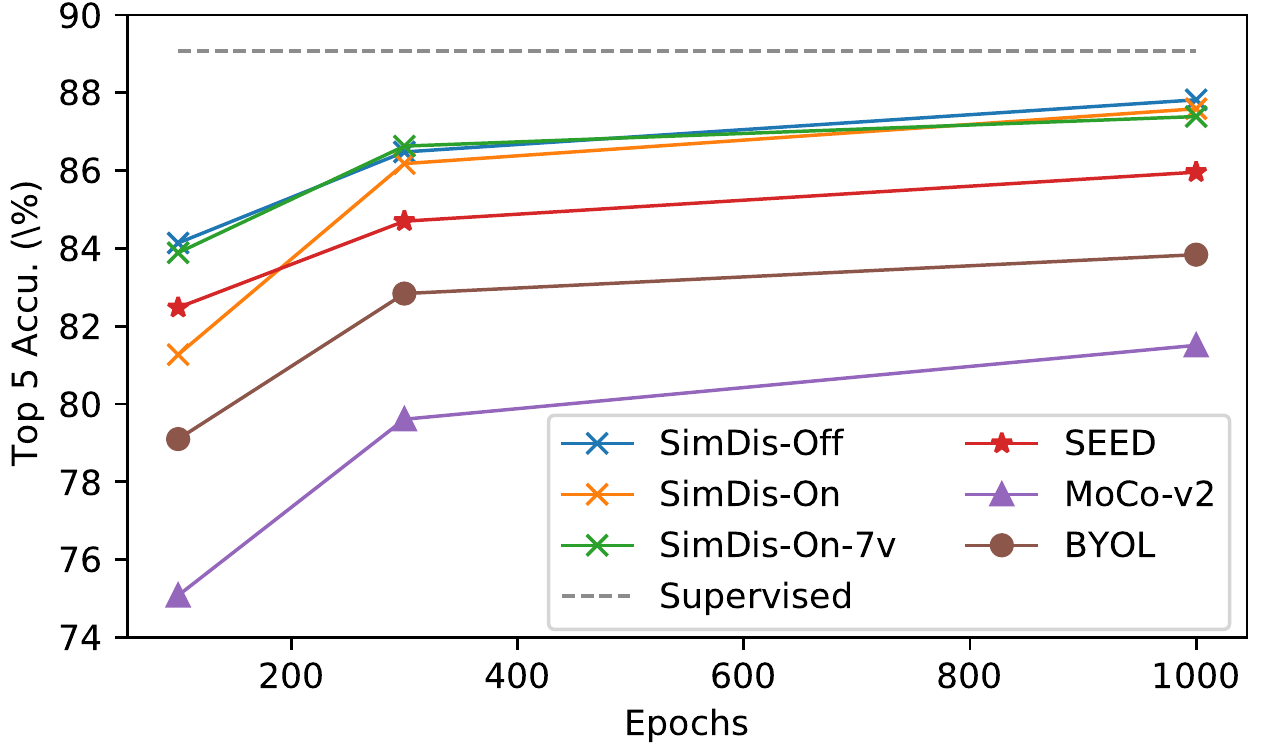}
        \caption{Top-5 Accuracy}
        \label{fig:top5}
    \end{subfigure} \hspace{0.1cm}
    \caption{The linear evaluation of ResNet18 backbone is reported. The self-supervised methods on ResNet18 show a large gap with the supervised counterpart. SEED boosts the ResNet18 backbone in the self-supervised setting. Both our online and offline methods improve the backbone further.}
    \label{fig:data_vis}
\end{figure*}

\section{Experiments}
In this section, we conduct three experiments to verify our baselines: 1) We demonstrate the pros and cons of online/offline distillation as well as our improvement; 2) We compare our baseline with other self-supervised model distillation methods; 3) We remove LARS and SyncBN for promoting a simple and practical setup.

\noindent\textbf{Architecture.} The teacher model is based on ResNet50, a commonly used network in the self-supervised setting. The ResNet50 backbone with an output dimension of 2048 is taken as the encoder in $T$. The projector is an MLP that consists of FC(4096)+BN+ReLU+FC(256). The predictor of $T$ uses the same architecture as the projector. The encoder and the projector of $\hat{T}$ have the same architecture as in $T$. The student model is constructed by replacing the ResNet50 with ResNet18.

\noindent\textbf{Optimization.} In the standard setting, the LARS optimizer with a cosine decay learning rate schedule and a warm-up period of 10 epochs is used. The base learning rate is set to 0.3 for 256 samples and linearly scaled up with the batch size. We use a batch size of 2048 split over 32 V100 GPUs, where the synchronized Batch Normalization (SyncBN) is applied. For $\hat{T}$ and $\hat{S}$, the exponential moving average parameter $\tau$ starts from 0.99 and is increased to one during training, following~\cite{grill2020bootstrap}. The models are trained on ImageNet-1k dataset and evaluated with the standard linear evaluation protocol~\cite{grill2020bootstrap}. The learning rate for linear evaluation is set as 0.001.

\subsection{Online vs. offline Distillation}
In this experiment, we compare the computational cost of online and offline distillation as well as their performance. The FLOPs of the models and their top-1 accuracy under different training epochs are reported in Tab.~\ref{tab:on_vs_off}.

The offline method requires more computational FLOPs than the online one. Compared with single-view distillation (\emph{SimDis-On}), the extra cost brought by Multi-view distillation (\emph{SimDis-On-7v}) is negligible. The offline baseline \emph{SimDis-Off} consistently outperforms the online baseline \emph{SimDis-On}, but the gap gradually closes with longer training. Though the gap is about 3.65\% for 100 epoch training, the multi-view distillation improves the online method by 2.57\%, narrowing the gap significantly with only negligible computational footprint.  We hypothesize that the multiple distillation targets provided by the teacher and the student make the gradient optimization direction more accurate. 

\begin{table*}[t]
\begin{center}
\begin{tabular}{ c c c |  c c c c |  c c c}
\toprule
\multicolumn{7}{c|}{Target views for prediction using $S$-v}  & \multirow{3}{*}{Num\_View} & \multirow{3}{*}{Online} & \multirow{3}{*}{Offline} \\
\cline{1-7} \\ [-1.em]
\multicolumn{3}{c|}{Views on Student}  & \multicolumn{4}{c|}{Views on Teacher}  &  &  &  \\
\cline{1-7} \\ [-1.em]
$S$-v{\tiny $^\prime$} &$\hat{S}$-v & $\hat{S}$-v{\tiny $^\prime$} & $T$-v & $T$-v{\tiny $^\prime$} &$\hat{T}$-v & $\hat{T}$-v{\tiny $^\prime$} &  &  &  \\
\midrule
& & \checkmark  &  & & &  & 1 & 55.96 & - \\
\hline
& & \checkmark  & \checkmark &  & & & 2& 58.20 & 59.67  \\
& & \checkmark  & &  \checkmark & &  &  2 & 58.29 & 62.08   \\
& & \checkmark  &  &  & \checkmark &  &2& 59.25 & 59.72 \\
& & \checkmark  & &  & & \checkmark &2& 58.84 & 62.66 \\
\hline
& & \checkmark  & \checkmark & \checkmark & & &3& 58.74 & 62.06 \\
& & \checkmark  &  & & \checkmark & \checkmark &3& 59.55 & \textbf{63.27} \\
\hline
& & \checkmark  & \checkmark  & \checkmark & \checkmark & \checkmark  &5& 59.06 & 59.12 \\
\hline
\checkmark & \checkmark & \checkmark  & \checkmark  & \checkmark & \checkmark & \checkmark & 7 & \textbf{63.52} & \textbf{63.14}\\
\bottomrule
\end{tabular}
\end{center}
\caption{The teacher models and the student models are trained for 100 epochs without syncBN or LARS. Different combinatons of the 7 representation views are explored. Generally, more views lead to the better performance.}
\label{tab:simple}
\end{table*}

\subsection{Self-supervised Model Distillation}
In this experiment, we compared our approach with various methods: ResNet18 trained with MoCo-v2 and BYOL, ResNet18 trained with SEED and our baselines. In SEED~\cite{fang2021seed}, we set the hyper-parameters as in~\cite{fang2021seed} and replace the teacher and the data processing correspondingly. We also provide supervised ResNet18 as an upper bound. We run Supervised, MoCo-v2 and SEED on a single node, following their settings. For the remaining methods, we follow the optimization setting described above.

Figure~\ref{fig:data_vis} reports the top-1 and top-5 accuracy with training epochs of 100, 300, and 1000. When ResNet18 is applied as the encoder, the self-supervised learning methods (i.e., BYOL and MoCo-v2) suffers a large performance gap with the supervised learning. The existing distillation method SEED can reduce the performance gap by a large margin. Our approach SimDis can further bridge the gap significantly. Both online and offline baselines achieve state-of-the-art performance. We hope they serve as strong baselines for the development of self-supervised distillation in the future.

\subsection{Removing LARS and SyncBN}
Our distillation approach is based on self-supervised learning. In current literatures, the self-supervised learning setting often requires LARS optimizer and SyncBN to stabilize the training process. However, the LARS optimizer and SyncBN are computationally expensive. 

We explore a simple and practical setting for self-supervised model distillation,  where none of LARS optimizer and SyncBN is applied and only a single machine with 8 GPUs is used. In this experiment, LARS optimizer and SyncBN are not applied, and the model is trained for 100 epochs with a batch size of 8*128 and a base learning rate of 0.05. We explore different variants of our approach and distill the student model with different teaching views and their combinations.

The performance is reported in Tab.~\ref{tab:simple}. The student model ResNet18 achieves 55.96 top1 accuracy when trained without distillation. In online setting, when one view from teacher is available, the student performance can be improved. In the case that all 7 views are available, the student achieves the best. In offline setting, a teacher with ResNet50 is first trained with BYOL approach for 100 epochs on the given single machine. The teacher achieves 65.83 top1 accuracy. Given the fixed teacher, the students are trained with different targets for 100 epochs. Since the teacher performs well even at the start of training, the students is improved with the help of teacher. When all 7 views are used, the online students are still comparable to the offline students.

\section{Conclusion}
In this work, we explore how to improve small self-supervised models with knowledge distillation. A strong baseline is proposed for future studies in this direction. Empirically experiments demonstrate that the proposed baseline achieves the state-of-the-art performance.

\bibliographystyle{ieee_fullname}
\bibliography{egbib}

\begin{thebibliography}{10}\itemsep=-1pt

\bibitem{ba2013deep}
Lei~Jimmy Ba and Rich Caruana.
\newblock Do deep nets really need to be deep?
\newblock {\em arXiv preprint arXiv:1312.6184}, 2013.

\bibitem{caron2020unsupervised}
Mathilde Caron, Ishan Misra, Julien Mairal, Priya Goyal, Piotr Bojanowski, and
  Armand Joulin.
\newblock Unsupervised learning of visual features by contrasting cluster
  assignments.
\newblock {\em arXiv preprint arXiv:2006.09882}, 2020.

\bibitem{chen2020simple}
Ting Chen, Simon Kornblith, Mohammad Norouzi, and Geoffrey Hinton.
\newblock A simple framework for contrastive learning of visual
  representations.
\newblock In {\em International conference on machine learning}, pages
  1597--1607. PMLR, 2020.

\bibitem{fang2021seed}
Zhiyuan Fang, Jianfeng Wang, Lijuan Wang, Lei Zhang, Yezhou Yang, and Zicheng
  Liu.
\newblock Seed: Self-supervised distillation for visual representation.
\newblock {\em arXiv preprint arXiv:2101.04731}, 2021.

\bibitem{grill2020bootstrap}
Jean-Bastien Grill, Florian Strub, Florent Altch{\'e}, Corentin Tallec,
  Pierre~H Richemond, Elena Buchatskaya, Carl Doersch, Bernardo~Avila Pires,
  Zhaohan~Daniel Guo, Mohammad~Gheshlaghi Azar, et~al.
\newblock Bootstrap your own latent: A new approach to self-supervised
  learning.
\newblock {\em arXiv preprint arXiv:2006.07733}, 2020.

\bibitem{gu2020search}
Jindong Gu and Volker Tresp.
\newblock Search for better students to learn distilled knowledge.
\newblock {\em arXiv preprint arXiv:2001.11612}, 2020.

\bibitem{he2020momentum}
Kaiming He, Haoqi Fan, Yuxin Wu, Saining Xie, and Ross Girshick.
\newblock Momentum contrast for unsupervised visual representation learning.
\newblock In {\em Proceedings of the IEEE/CVF Conference on Computer Vision and
  Pattern Recognition}, pages 9729--9738, 2020.

\bibitem{hinton2015distilling}
Geoffrey Hinton, Oriol Vinyals, and Jeff Dean.
\newblock Distilling the knowledge in a neural network.
\newblock {\em arXiv preprint arXiv:1503.02531}, 2015.

\bibitem{koohpayegani2020compress}
Soroush~Abbasi Koohpayegani, Ajinkya Tejankar, and Hamed Pirsiavash.
\newblock Compress: Self-supervised learning by compressing representations.
\newblock {\em arXiv preprint arXiv:2010.14713}, 2020.

\bibitem{oord2018representation}
Aaron van~den Oord, Yazhe Li, and Oriol Vinyals.
\newblock Representation learning with contrastive predictive coding.
\newblock {\em arXiv:1807.03748}, 2018.

\bibitem{romero2014fitnets}
Adriana Romero, Nicolas Ballas, Samira~Ebrahimi Kahou, Antoine Chassang, Carlo
  Gatta, and Yoshua Bengio.
\newblock Fitnets: Hints for thin deep nets.
\newblock {\em arXiv preprint arXiv:1412.6550}, 2014.

\bibitem{tian2019contrastive}
Yonglong Tian, Dilip Krishnan, and Phillip Isola.
\newblock Contrastive multiview coding.
\newblock {\em arXiv preprint arXiv:1906.05849}, 2019.

\bibitem{wu2018unsupervised}
Zhirong Wu, Yuanjun Xiong, Stella~X Yu, and Dahua Lin.
\newblock Unsupervised feature learning via non-parametric instance
  discrimination.
\newblock In {\em Proceedings of the IEEE Conference on Computer Vision and
  Pattern Recognition}, pages 3733--3742, 2018.

\bibitem{you2017learning}
Shan You, Chang Xu, Chao Xu, and Dacheng Tao.
\newblock Learning from multiple teacher networks.
\newblock In {\em Proceedings of the 23rd ACM SIGKDD International Conference
  on Knowledge Discovery and Data Mining}, pages 1285--1294, 2017.

\bibitem{zhang2018deep}
Ying Zhang, Tao Xiang, Timothy~M Hospedales, and Huchuan Lu.
\newblock Deep mutual learning.
\newblock In {\em Proceedings of the IEEE Conference on Computer Vision and
  Pattern Recognition}, pages 4320--4328, 2018.

\end{thebibliography}

\end{document}